\documentclass[letterpaper, 10 pt, journal, twoside]{IEEEtran}
\usepackage{graphicx}
\usepackage[table,xcdraw]{xcolor}
\usepackage{amsmath}
\usepackage{amssymb}
\usepackage{booktabs}
\usepackage{multirow}
\usepackage{color}
\usepackage{adjustbox}
\usepackage{xcolor}
\usepackage{cite}
\usepackage{algorithm}
\usepackage[vlined, ruled, algo2e]{algorithm2e}
\usepackage{caption}
\usepackage{subcaption}
\usepackage{csquotes}
\usepackage{dblfloatfix}   
\usepackage{float}
\usepackage[flushleft]{threeparttable}

\usepackage{varwidth}
\ifCLASSINFOpdf
\else
\fi
\renewcommand{\Indentp}[1]{%
	\advance\leftskip by #1
	\advance\skiptext by -#1
	\advance\skiprule by #1}%
\renewcommand{\Indp}{\algocf@adjustskipindent\Indentp{\algoskipindent}}
\renewcommand{\Indm}{\algocf@adjustskipindent\Indentp{-\algoskipindent}}
\makeatother

\ifodd 1

\else
\fi

\begin{document}
%
\title{Multi-AGV's Temporal Memory-based RRT Exploration in Unknown Environment}

%
%
%

\author{Billy~Pik~Lik~Lau, Brandon~Jin~Yang~Ong, Leonard~Kin~Yung~Loh, Ran~Liu, Chau~Yuen,  Gim~Song~Soh, and U-Xuan~Tan%
\thanks{Manuscript received: Feb, 23, 2022; Revised Jun, 22, 2022; Accepted July, 15, 2022.}
\thanks{All authors are with Engineering Product Development, Singapore University of Technology and Design, Singapore. (Corresponding Author: Billy Pik Lik Lau, Email: {\tt \small billy\_lau@sutd.edu.sg})}
\thanks{Source Code: https://github.com/hikashi/TM-RRT\_exploration}
\thanks{Video: https://youtu.be/PbUAtM8wZdc}
\thanks{Digital Object Identifier (DOI): 10.1109/LRA.2022.3190628}
}
%
%

\markboth{Paper Accepted 15th July 2022, published at IEEE Robotics and Automation Letters, DOI: 10.1109/LRA.2022.3190628}
{Lau \MakeLowercase{\textit{et al.}}: Multi-AGV's TM-RRT Exploration in Unknown Environment} 

%



\maketitle


\begin{abstract}
With the increasing need for multi-robot for exploring the unknown region in a challenging environment, efficient collaborative exploration strategies are needed for achieving such feat.
A frontier-based Rapidly-Exploring Random Tree (RRT) exploration can be deployed to explore an unknown environment.
However, its' greedy behavior causes multiple robots to explore the region with the highest revenue, which leads to massive overlapping in exploration process.
To address this issue, we present a temporal memory-based RRT (TM-RRT) exploration strategy for multi-robot to perform robust exploration in an unknown environment.
It computes adaptive duration for each frontier assigned and calculates the frontier's revenue based on the relative position of each robot.
In addition, each robot is equipped with a memory consisting of frontier assigned and share among fleets to prevent repeating assignment of same frontier.
Through both simulation and actual deployment, we have shown the robustness of TM-RRT exploration strategy by completing the exploration in a $25.0m\times54.0m$ ($1350.0m^2$) area, while the conventional RRT exploration strategy falls short.
\end{abstract}


%
\IEEEpeerreviewmaketitle

\section{Introduction}
\label{sec:Introduction}

\IEEEPARstart{M}{ulti-robot} exploration using an autonomous ground vehicle (AGV) is a research topic that has received increasing interest from a lot of research communities as well as the industry's big players.
Such technology can be used to explore an unknown region in the scenario such as collapsed buildings, unexplored caves, tunnels, etc.
This helps to reduce the chances of risking human personnel when needed. 
With the advancement of the hardware and sensor fusion techniques~\cite{Lau2019survey, yassin2016recent}, a fleet of AVG can be coordinated to explore the unknown region, outperforming single AVG exploration.
To date, there are a few generic exploration strategies in the literature such as frontier-based~\cite{yamauchi1998frontier,Tian2020Autonomous}, information-based~\cite{stachniss2005information,Liu2021Prior}, machine learning-based~\cite{Niroui2019Deep}, and hierarchical-based~\cite{cao2021tare,Cao2021Exploring}. 
A more detailed comparison of each exploration strategy can be found in~\cite{QuattriniLi2020Exploration, Amigoni2017Multirobot}.
Among them, the frontier-based exploration strategy is easier to implement due to separate modules from move base and localization.

Frontier-based exploration consists of two major processes, which are frontier detection~\cite{Yamauchi1997frontier,yamauchi1998frontier} and tasks assignment~\cite{Williams2017Decentralized}.
First, the detection of the unknown region (which is also known as a frontier) for the map relies on the sensor types and algorithm used.
Examples of sensor used for exploration include visionary~\cite{Fraundorfer2012Vision, Andre2014Coordinated}, LiDAR\cite{Umari2017Autonomous,Niroui2019Deepa}, and camera~\cite{Santosh2008Autonomous}.
Secondly, another core element of multi-robot exploration depends on the tasks allocation and the common examples of task assignment strategy are: probabilistic~\cite{Zhang2020Rapidly}, machine learning~\cite{Niroui2019Deepa}, auction or market-based~\cite{siciliano2008springer, Nam2015Assignment,zlot2006market}, etc.

The challenges of multi-AGVs exploration can be broken down into the following lists: 
First, multiple AGVs need to carry out exploration missions in a robust manner.
The proposal of designing robust multi-robot planning can be found in these works~\cite{madridano2021trajectory, rekleitis2001multi}. 
However, seamless multi-robot exploration still requires more works in real-world applications with different types of constraints.
Secondly, the goal assignment of the exploration ought to consider each other AGV's information for better exploration coordination.
As discussed earlier in task assignments, the tasks assignment plays an important role on determining exploration effiency.
Therefore, optimizing the tradeoff between communication overhead, goal assignment, information gain of each frontier, and number of robots is crucial for efficient multi-robot exploration.
Thirdly, the exploration algorithm needs to be robust and able to perform exploration regardless of the environment surface area especially large unknown environment.
Tian et. al~\cite{Tian2020Autonomous} has studied exploration for area size of $184m^2$, where Umari et.al~\cite{Umari2017Autonomous} have performed exploration at area size of $182m^2$.
Meanwhile, a larger 3D map exploration has been studied in~\cite{Cao2021Exploring} covering up to $320m\times320m\times120m$ but only limited to single robot exploration.
Currently, large map exploration over $1000m^2$ mostly are studied by research groups participating in DARPA challenges and these works can be found in~\cite{Dang2020Graph, Tranzatto2022CERBERUS, Agha2021Nebula}.
\begin{figure}[hb]
	\centering
	\vspace{-0.1cm}
	\includegraphics[width=0.43\textwidth]{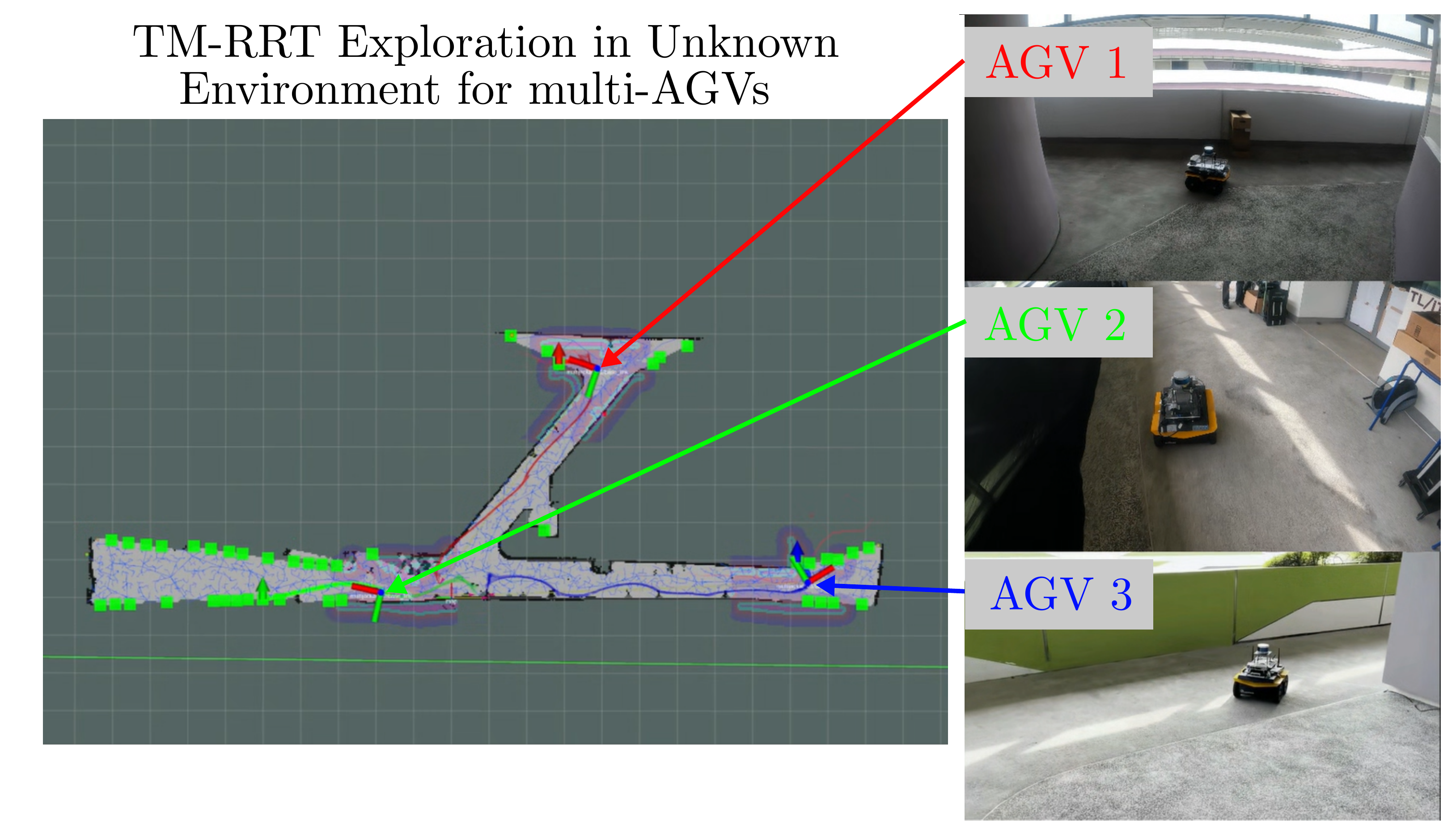} 
	\vspace{-0.2cm}
	\caption{Example of TM-RRT exploration deployed in three AGVs.}
	\label{fig:simulationExample}
	\vspace{-0.46cm}
\end{figure}

\begin{figure*}[hb]
	\centering
	\vspace{-0.2cm}
	\includegraphics[width=0.719\textwidth]{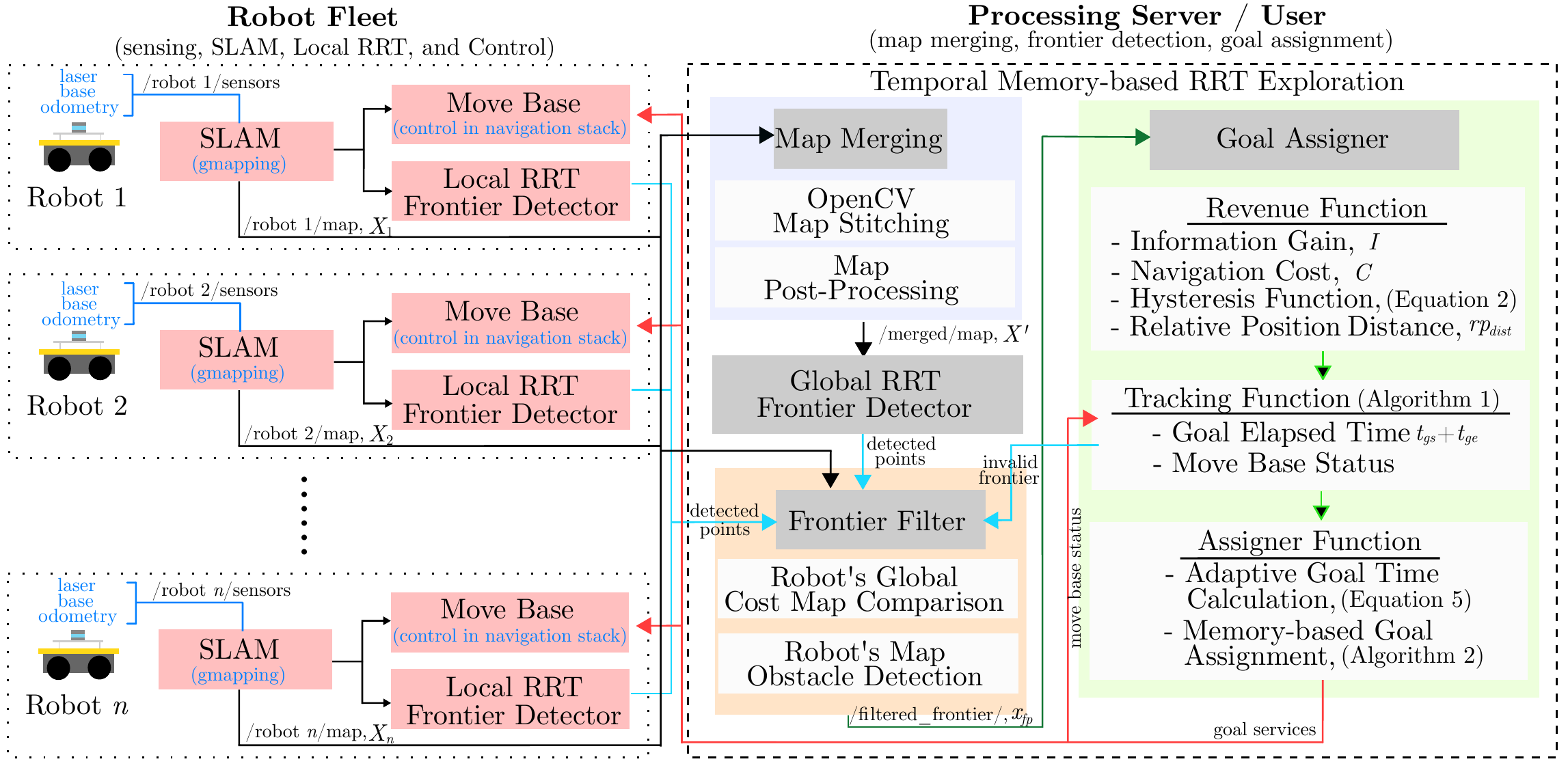} 
	\vspace{-0.2cm}
	\caption{System architecture for multi-robot TM-RRT exploration.}
	\label{fig:system_architecture}
	\vspace{-0.16cm}
\end{figure*}

With these challenges in mind, we propose a Temporal Memory-based Rapidly-Exploring Random Tree (TM-RRT) exploration for multi-AGV in an unknown environment.
It utilizes the concept of temporal-based goal assignment~\cite{Nikou2020Scalable} to set a time limit for an AGV to explore a given frontier while keeping track of past assigned goals.
This averts the AGV from using excessive time exploring the same frontier or being stuck in an unreachable frontier.
Another important element of the TM-RRT exploration is to keep track of the AGV's goal history (memory module), which avoids the AGV from having repeated frontier being assigned.
To further improve the goal assignment, the revenue function includes the AGV's relative position when it calculates the revenue for each given frontier. 
In short, the contributions of this paper are listed as follow:
\begin{itemize}
	\item We proposed a temporal and memory-based frontier exploration, which governs the duration of frontier assigned to an AGV. This mitigates the time spent for an AGV reaching a certain frontier, while it is unable to obtain a ``goal reached'' state. The memory part prevents the assignment of identical goals to AGVs.
	\item We introduced a relative distance metric of each AGV when calculating the revenue for each given frontier. This allows an AGV to keep track of each other relative position and therefore reducing ``multiple AGV end up exploring frontier in close proximity'' behavior.
	\item We proposed a map post-processing method that helps with reducing localization noises from larger LiDAR scanning range, which later used for frontier detection.
	\item We implemented and demonstrated TM-RRT exploration in an actual multi-AGV deployment in an unknown environment size of $1350m^2$ ($25.0m\times54.0m$), where conventional RRT exploration is having difficulty performing efficiently.
\end{itemize}
Through the simulation, it shows an improvement over the existing RRT exploration method around $30\%$ in exploration duration and $50\%$ in traveled distance over two different maps ($120.00m^2$ and $376.25m^2$).
Moreover, we have implemented the TM-RRT exploration in a physical implementation is shown in Fig.~\ref{fig:simulationExample}.
Note that green dot in the map represents frontier, while the traveled path of each AGV is denoted by red, green, and blue lines respectively.

The remainder of this paper is organized in the following order: 
First, in Section~\ref{sec:architecture}, we define the preliminary terminology and introduce the system architecture of the Robot Operating System (ROS) network. 
In addition, we also illustrate the multi-ROS master setup used for the real-world experiment.
Next, the working mechanism of the proposed TM-RRT assignment module is explained in Section~\ref{sec:TMrrt}. 
We focus on two core modules, which are map-merging modules and goal-assigner modules.
The former cleans up the map while the latter consists of the adaptive duration function computation and memory-based frontier assignment.
Subsequently, simulation is carried out in Section~\ref{sec:simulation} to measure TM-RRT exploration's performance in different parameters and different map complexity.
Subsequently in Section~\ref{sec:deploymnet}, TM-RRT exploration is implemented with three AGVs in an unknown environment, and observation about its performance is analysed. 
Lastly, we conclude our work in TM-RRT exploration in Section~\ref{sec:conclusion} as well as discuss possible future improvements.

\section{Overview of Temporal Memory-based RRT Exploration Strategy}
\label{sec:architecture}

In this section, we describe the overview of the TM-RRT exploration strategy through four different subsections.
First, we define the terminology used for describing modules used in the TM-RRT exploration.
Next, we present the system architecture of the TM-RRT exploration and briefly explain the working flow.
Lastly, we talk about the implementation of ROS multi-master in TM-RRT exploration.


\subsection{Preliminary terminology}
\label{subsec:preliminary_term}

The terminology definitions usd in the TM-RRT exploration are presented as follows:
\begin{itemize}
	\item \textbf{Map} $X$: A 2D matrix that consists set of total space, which consists of occupied (obstacles), unoccupied (free space), and unknown (unexplored space). Each of them is represented by $100$, $0$, and $-1$ respectively.
	\item \textbf{Occupancy grid}: A 2D representation of the maps that divide the map into cells, which is indexed by its coordinates. Similarly, each cell value is represented by a set of values ${0,1,-1}$ that denote free, occupied, and unknown regions.
	\item \textbf{Free Space} $X_{free}$: A set of values that denote a region that has been explored and not occupied by obstacles.
	\item \textbf{Unknown Region}: The unknown region denotes the subset of the map area that has not been explored.
	\item \textbf{Navigation Cost} $C$: The expected cost (heuristic distance) for an AGV to traverse from the current position to a given frontier.
	\item \textbf{Information Gain} $I$: The area of the unknown region given a coordinate for an AGV to explore governed by a radius.
	\item \textbf{Relative Position Distance} $rp_{dist}$: The vector list of the distance between each AGV's current position and a frontier. This also can be used to calculate the distance between each assigned frontier and AGV's current position.
	\item \textbf{Frontier} $x_{f p}$: Clustered set of the unknown area using mean shift clustering~\cite{Cheng1995Mean} that reduce the point of unknown region to be explored.
	\item \textbf{Filtered Frontier}: List of frontiers that has been filtered given the following conditions - region has been explored and frontier that is marked as invalid.
	\item \textbf{Invalid Frontier}: Frontiers that is not reachable given the following conditions - move base return an error, the frontier is located inside the obstacle, and the duration taken to explore a given frontier is too long.
	\item \textbf{Goal Assigned List} $x_{G_{\{1,...,n\}}}$: List of frontiers that has been previously assigned to the AGV. Note that, $G_{\{1,...,n\}}$ represents the goal assigned to AGV $\{1,...,n\}$.
	\item \textbf{Goal Start Time} $t_{gs}$: The time when a frontier goal $x_{fp}$ is being assigned to an AGV.
	\item \textbf{Goal Expected Time} $t_{ge}$: The expected duration for an AGV to reach a frontier goal $x_{fp}$.
\end{itemize}
\vspace{-0.2cm}

\subsection{System architecture}
\label{subsec:systemarchitect}
In this subsection, we detail the architecture proposed for multi-AGV TM-RRT exploration, which leverages the centralized approach.
A centralised approach requires a server to process the information and coordinate between a fleet of AGVs. 
An illustration of the architecture is presented in Fig.~\ref{fig:system_architecture}.
The strategy deployed for the TM-RRT exploration can be segregated into four parts, which are map merging, frontier detector, frontier filter, and frontier assigner.

In our approach, the frontier detector detects the unexplored region (known as detected points) using the merged map as input with the RRT approach~\cite{Umari2017Autonomous} locally at AGV and globally at server.
The frontier detector module is implemented locally at the AGV (robot fleet) and the control station (processing server).
Next, the detected points are fed into the filter module to perform clustering on the frontier that is within proximity. 
This step gradually reduces the number of points and cluster them into centroids for the assigner module to process.
Using the filtered frontier, the assigner module allocates the frontier to the AGV to epxlore based on the highest revenue. 
To calculate revenue for each given frontier, we need to consider following information, which are information cost, navigation cost, and relative position. 
Moreover, an adaptive goal is computed for the assigned frontier to ensure AGV completes the given frontier within the time frame.
Lastly, to maintain the operability of the TM-RRT exploration, a monitoring module is deployed alongside to ensure AGV executes the mission within the given time or does not stuck (move base reporting error message due to certain problem, e.g. oscillation).

\subsection{ROS multi-master setup}
As for real-world multi-AGV deployment, we use multi-ROS master architecture~\cite{juan2015multi} to manage the multi-robot exploration task involving multiple AGVs running individual ROS navigation stacks.
The Multi-ROS Master to allow interoperability between different ROS Master for each AGV.
This allowed for some advantages when compared to a single ROS master setup:
\begin{itemize}
	\item Individual AGVs could have their own ROS networks with identical topic names, with no risk of topic name clashes, allowing for AGV software to be duplicated with minimal changes.
	\item A subset of topics could be exposed to the shared Multimaster network having greater control over what topics are shared, preventing name clashes, and allowing for certain topics to be made available to certain hosts (e.g., AGVs cannot subscribe to each other, but the server can subscribe to all AGV hosts' topic).
	\item It allows certain topics to be ignored or hidden, minimising network traffic.
	\item Individual AGV nodes could continue to function even when disconnected from the shared network, since they would still be connected to their local ROS master, allowing for uninterrupted conduct of navigation.
\end{itemize}

\begin{figure}[h]
	\centering
	\vspace{-0.2cm}
	\includegraphics[width=0.48\textwidth]{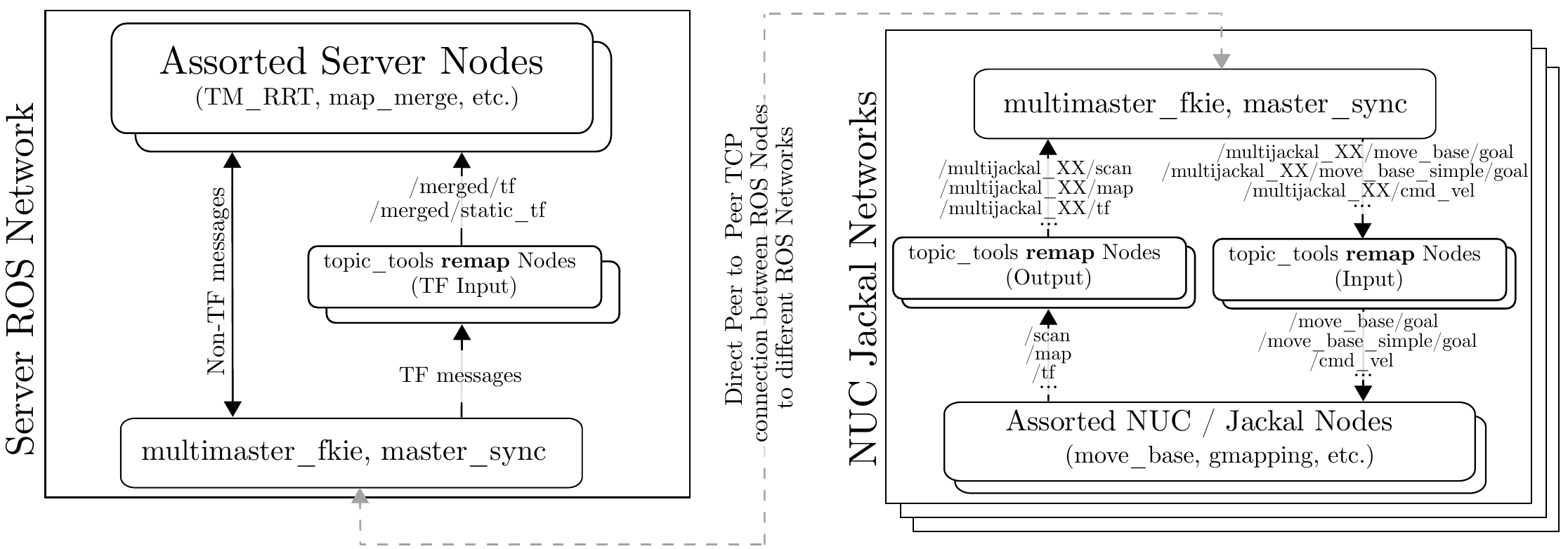} 
	\vspace{-0.2cm}
	\caption{ROS network and multimaster architecture for three jackals implementing TM-RRT exploration.}
	\label{fig:ros_multimaster}
	\vspace{-0.26cm}
\end{figure}

This configuration allows us to work with multiple compute units using a common network as well as providing ROS topic control to govern which topic to broadcast.
In the experiment setup, nodes were spread out across multiple ethernet and WiFi subnets. 
Network routing rules are implemented to ensure only connected ROS nodes are reachable within the same network.
An illustration of the network setup between server and Jackal is illustrated in Fig.~\ref{fig:ros_multimaster}.

Since each AGV has its own ROS master, no explicit topic name spacing was done at the AGV level.
During communication with the server, topic name spacing was done through ``topic\_tools'' to prepend the topic name with the AGV's name space.
Furthermore, for TransForm(TF)~\cite{Foote2013tf} messages to be aggregated, the individual robot TF topics were all remapped onto a single TF topic for server processing.

\section{Temporal Memory-based RRT Exploration}
\label{sec:TMrrt}

In this section, we describe the alteration of RRT exploration based on the modules introduced earlier in Section~\ref{sec:architecture}. 
The TM-RRT exploration can be divided into four parts, which are the map merging module, RRT frontier detector module, frontier filtering module, and goal assigner module. 
The RRT frontier detector module is responsible for detecting the unknown region based on the local map or merged map.
Afterwards, the filtering module is used to filter detected frontier points that are either obstacle, explored space, or the unexplored region that does not yield high information gain.
These two modules use the same function as the original RRT exploration~\cite{Umari2017Autonomous}.
Therefore, the focus of our work is to improve two modules, which are (1) map-merging module and (2) temporal memory-based goal assigner module.

\subsection{Map-merging module}
\begin{figure}[h]
	\centering
	\vspace{-0.2cm}
	\includegraphics[width=0.44\textwidth]{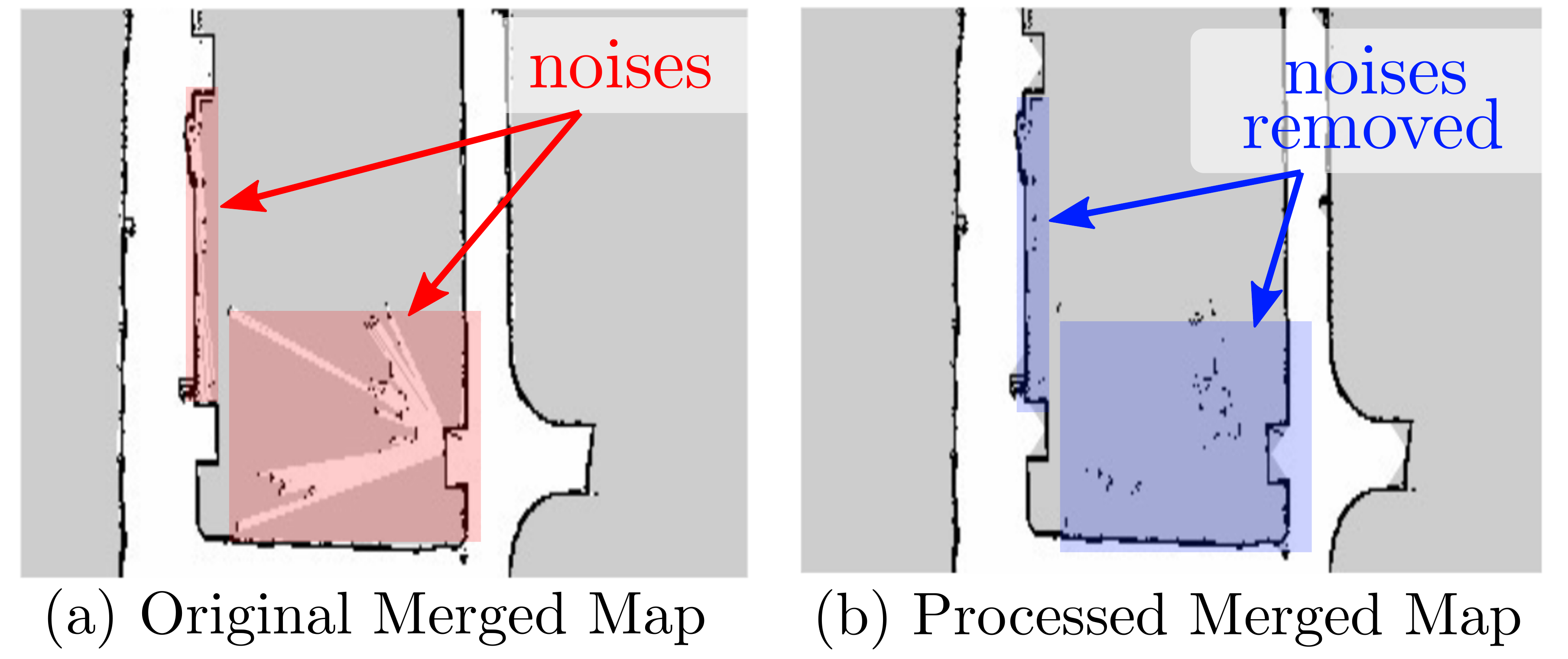} 
	\vspace{-0.2cm}
	\caption{Comparison of the original merged map and processed merged map.}
	\label{fig:map_preprocess}
	\vspace{-0.16cm}
\end{figure}

The map-merging module is crucial for combining individual maps of each AGV so that it can be used to identify the unexplored region.
To generate merged-map $X$ in TM-RRT exploration, we utilize multi-robot map merge~\cite{horner2016map} to combine local maps $X_{\{1,...,n\}}$ from individual AGV.
In the map merging process for experiment, it suffers from noises from the localization module due to the larger LiDAR scanning range (range of $20.0m$ - $30.0m$) as shown in Fig.~\ref{fig:map_preprocess}(a).
This causes a problem such as the tree of RRT expanding to the area that is supposed to be an obstacle or unreachable (TF drifting or RRT expands inside the obstarcle), which leads to stuck AGV during exploration.

To tackle this issue, a map post-processing step is proposed, where the free regions of the map (white) were isolated and image morphological operations were applied to remove the less densely connected spurious-free regions. 
Specifically, 6 iterations of image erosion with an elliptical kernel of $3$-by-$3$, followed by $6$ iterations of image dilation with an elliptical kernel of $3$-by-$3$ were applied to the free regions of the merged map.
The result is a cleaner map as illustrated in Fig.~\ref{fig:map_preprocess}(b) with ambiguous free region removed along with a large buffer of marked unknown space, which makes it ideal for frontier detection.
As for the deployment, we include the merged map into the frontier filtering module and examine whether a frontier belongs to a free, occupied or unknown space.
This is to ensure the frontier filtering module follows the merged map generated and filters out the known region.
\vspace{-0.2cm}

\subsection{Temporal Memory-based RRT Goal Assigner Module}
In this subsection, we depict the working mechanism of the TM-RRT exploration's assignment module.
The working mechanism can be split into two steps, which are (1) revenue function with relative position and (2) temporal memory-based goal assignment.

\subsubsection{Revenue function with relative position, $rp_{dist}$}
\label{subsec:revenue_function}

After utilizing RRT to perform the detection point of the unexplored region, we perform the mean-shift clustering to group the detected point into one centroid (filtered frontier) as shown in Fig.~\ref{fig:system_architecture}. 
Upon receiving filtered frontiers, the assigner will perform calculation of the revenue function for each AGV given the filtered frontiers.

To mitigate multiple AGVs having same revenue gain given the same filtered frontier, we rewrite Equation~\ref{eqn:revenue_function}, revenue function $R$ of a given frontier point $x_fp$:
\begin{equation}
\small
\begin{aligned}
R\left(x_{f p}\right) &=\left(  \lambda h\left(x_{f p}, x_{r}\right) I\left(x_{f p}\right)f(x_{f p}, x_{G_{\{1,...,n\}}}) \right) -C\left(x_{f p}\right), 
\end{aligned}
\label{eqn:revenue_function}
\end{equation}
where $x_r$ denotes AGV's current position $x_r$ and $\lambda$ is a constant weight for adjusting information gain, $I(x_{fp})$ and relative distance function, $f(x_{f p},x_{G_{\{1,...,n\}}})$ weightage.
$x_{G_{\{1,...,n\}}}$ denotes list of AGV's current goal at the current timestamp, while $x_{f p}$ represents the frontier that are being considered to be designated as goal.
The $C\left(x_{f p}\right)$ represents the navigation cost of AGV to traverse from current position to the given frontier, $x_{f p}$. 

Next, we compute the hysteresis gain~\cite{simmons2000coordination}, $ h\left(x_{f p}, x_{r}\right)$ as Equation~\ref{eqn:hysteresis_gain}:
\begin{equation}
\begin{aligned}
h\left(x_{f p}, x_{r}\right) &= \begin{cases}1.0, & \text { if }\left\|x_{r}-x_{f p}\right\|>h_{\text {rad }} \\
h_{\text {gain }}, & h_{\text {gain }}>1.0\end{cases},
\end{aligned}
\label{eqn:hysteresis_gain}
\end{equation}
where it calculates the distance for an AGV current location $x_r$ and frontier $x_{f p}$.
Note that this can be adjusted based on the user preference. 
The main purpose of such function is to ensure AGV has better revenue within its $h_{rad}$ distance.

Subsequently, the relative distance function, $f(x_{f p}, x_{G})$ is introduced to compute the relative position of each AGV's goal $x_{G_{\{1,...,n\}}}$ and the frontier given $x_{f p}$.
Equation~\ref{eqn:relativeDistance} follows show the calculation of the relative distance function:
\begin{equation}
\scriptsize
\begin{aligned}
f(x_{f p}, x_{G}) = \begin{cases}
0.01 & \frac{min( || x_{G}-x_{f p}|| )}{rp_{dist}} < 0.01 \\
\frac{min(||x_{G}-x_{f p}||)}{rp_{dist}}   & 0.01 \ge \frac{max(|| x_{G}-x_{f p}||)}{rp_{dist}} \le 1.00 \\
1.00 & \frac{min( || x_{G}-x_{f p}|| )}{rp_{dist}} > 1.00
\end{cases},
\end{aligned}
\label{eqn:relativeDistance}
\end{equation}
where $x_{G_{\{1,...,n\}}}$ denotes the position of other AGV's current goal (currently assigned frontier) and $rp_{dist}$ is the distance threshold for calculating the nearest relative distance of AGV to a given frontier.
The function $min( || x_{G}-x_{f p}|| )$ compute the closest distance of the relative distance given the frontier and list of each AGV's current goal position.
This function aims to make sure each AGV is located far away from each other so that revenue calculation for the given frontier $x_{g p}$ stays at the highest value.

\subsubsection{Temporal memory-based goal assignment} 
\label{subsec:tm-rrt-assigner}

In this subsection, we explain the working mechanism of the memory-based goal assignment for the TM-RRT exploration.
It consists of two parts, which are adaptive duration assignment for frontier and memory-based frontier assignment.

To prevent an AGV from working on a goal for too long, we implement a temporal-based goal assignment for each frontier.
Each goal (frontier with the highest revenue) is assigned an adaptive duration depending on the distance between AGV's position and frontier.
The adaptive duration (duration in seconds) for each goal is calculated using the following Equation~\ref{eqn:adaptiveTime}:
\begin{equation}
\small
\begin{aligned}
k(x_{f p}, x_{r},z) = \begin{cases}
t_{pm} & ||x_{r} - x_{f p}|| < h_{rad} \\
t_{pm} \times ||x_{r} - x_{f p}||   & h_{rad} \ge ||x_{r} - x_{f p}|| \le z \\
t_{pm} \times z & ||x_{r} - x_{f p}|| > z
\end{cases},
\end{aligned}
\label{eqn:adaptiveTime}
\end{equation}
where $t_{p m}$ represents the expected average velocity and $z$ denotes the distance threshold for $t_{p m}$.
It requires a fine-tuning between the $z$ value and $t_{p m}$ to govern the duration assigned for an AGV to explore frontier goal.
For each goal assigned, expected duration given $t_{g e}$ can be computed using the following Equation~\ref{eqn:expectedTime}:
\begin{equation}
t_{g e, x_{f p}} = t_{g s} + k(x_{f p}, x_{r},z),
\label{eqn:expectedTime}
\end{equation}
where $t_{g s}$ denotes the current time a frontier being assigned to an AGV.

\begin{algorithm}[ht]	
	\caption{Temporal and Movebase Tracking Algorithm}
	\label{alg:monitorTemporal}
	\fontsize{8pt}{8pt}\selectfont
	\KwData{\textit{expected\_duration }$t_{g e}$, \text{robot\_movebase}}
	\KwResult{\textit{invalid\_frontier\_list}, }
	\BlankLine 
	\textbf{function} MonitorStateAndDuration() \BlankLine \vspace{-0.11cm}
	1. Perform tracking if the ROS node is active:\BlankLine \vspace{-0.11cm}
	\While{ROS not shut\_down}{
		\For{each AGV in robot\_namelist}{
			Check the expected duration: \BlankLine \vspace{-0.11cm} 
			\If{ $(t_{g e} - \text{now} )\le 0$}{
				add current $x_{f p}$ into the invalid frontier list \BlankLine \vspace{-0.11cm}
				cancel current goal and set AGV to idle \BlankLine \vspace{-0.11cm}}
			Check move base status: \BlankLine \vspace{-0.11cm} 
			\If{move base status return error}{
				add current $x_{f p}$ into the invalid frontier list \BlankLine \vspace{-0.11cm}
				cancel current goal and set AGV to idle \BlankLine \vspace{-0.11cm}}
	}}
	2. Publish the invalid frontier list for visualization\BlankLine \vspace{-0.11cm}
\end{algorithm}

Next, we implement a monitoring algorithm to ensure the frontier assigned to AGV would not take longer than the expected duration $t_{g e}$. 
The monitoring process is described in Algorithm~\ref{alg:monitorTemporal}.
In general, the algorithm keeps track of the assigned goal duration and the move base status for each AGV.
Should the AGV exceed the allocated duration, the algorithm will cancel the assigned goal and put AGV into an idle state.
For the move base status monitoring, we have referred to the move base messages defined in~\cite{guimaraes2016ros} to check for its statuses.
This ensures the AGV will label the unreachable frontier as an invalid frontier as soon as possible, and move on to the next highest revenue frontier. 

Subsequently, we deploy temporal memory-based goal assignment as shown in Algorithm~\ref{alg:rrtAssign}. 
\begin{algorithm}[h]
	\caption{Temporal Memory-based Goal Assigner}
	\label{alg:rrtAssign}
	\fontsize{8pt}{8pt}\selectfont
	\KwData{\textit{revenue\_list, robot\_namelist}}
	\BlankLine 
	\textbf{function} TM-RRT\_Assigner()  \BlankLine \vspace{-0.11cm}
	While there is unexplored region in the map:\BlankLine \vspace{-0.11cm} 
	\While{len(frontier\_list)$>0$}{
		Shuffle AGV sequence in robot\_namelist \BlankLine \vspace{-0.11cm}
		\For{AGV in shuffled\_robot\_namelist}{
			skip\_Assign = False \BlankLine \vspace{-0.11cm}
			not\_Assign  = True \BlankLine \vspace{-0.11cm}
			attempt\_count = $0$ \BlankLine \vspace{-0.11cm}
			\If{robot\_state is Busy}{
				$dist = ||x_{r p} - x_{g'}||$ \BlankLine \vspace{-0.11cm}
				\If{$dist > (1.5 \times h_{rad})$ and $dist \le (1.5\times rp_{dist})$}{skip\_Assign = True}
			}
			\If{not skip\_Assign}{a
				\While{(not\_Assign and attempt\_count) $<$ len(frontier\_list)}{
					$goal$  = getHighestRevenue(attempt\_count, revenue\_list) \BlankLine \vspace{-0.11cm}
					flag1 = invalid\_frontier\_check($goal$) \BlankLine \vspace{-0.11cm}
					flag2 = check\_own\_memory($goal, x_{g}$) \BlankLine \vspace{-0.11cm}
					flag3 = check\_other\_memory($goal, x_{g}$) \BlankLine \vspace{-0.11cm}
					\If{flag1 and flag2 and flag3}{ 
						1. calculate duration using Equation~\ref{eqn:adaptiveTime} \BlankLine \vspace{-0.11cm}
						2. update AGV history \BlankLine \vspace{-0.11cm} 
						3. send goal to AGV \BlankLine \vspace{-0.11cm}
					}
					\Else{attempt\_count $+= 1$}
				}
			}
	}}
\end{algorithm}
After calculating the revenue for each frontier, the TM-RRT assigner randomizes the AGV sequence to allow each AGV to get an opportunity to be assigned a goal before others. 
Each AGV's current state will be checked next before determining it should forfeit the current goal and move on to the next goal.

If the AGV's current state is busy, the assigner will have to decide whether to interrupt the AGV's current mission with a new frontier.
The condition for interruption is based on the distance of the AGV's current position and the goal, which are distances lesser than $1.5\times h_{rad}$ or greater than $1.5 \times rp_{dist}$.
The idea here is to enable interruption only when the AGV is either very close to the goal or too far away from the goal.
The optimality of the interruption value still requires further investigation.
Subsequently, the TM-RRT assigner will check for three conditions before assigning a new goal for idling AGVs or busy AGV that match the interruption condition. 
Firstly, the frontier with the highest revenue will undergo comparison against the invalid frontier list to make sure the invalid frontier is not repeated. 
Secondly, the frontier will be examined against its own existing goal history to make sure the appointed goal is not repeated. 
Lastly, the same process is repeated with other AGV's goals to ensure the frontier has not been previously assigned to other AGV. 
\begin{figure}[h]
	\centering
	\includegraphics[width=0.495\textwidth]{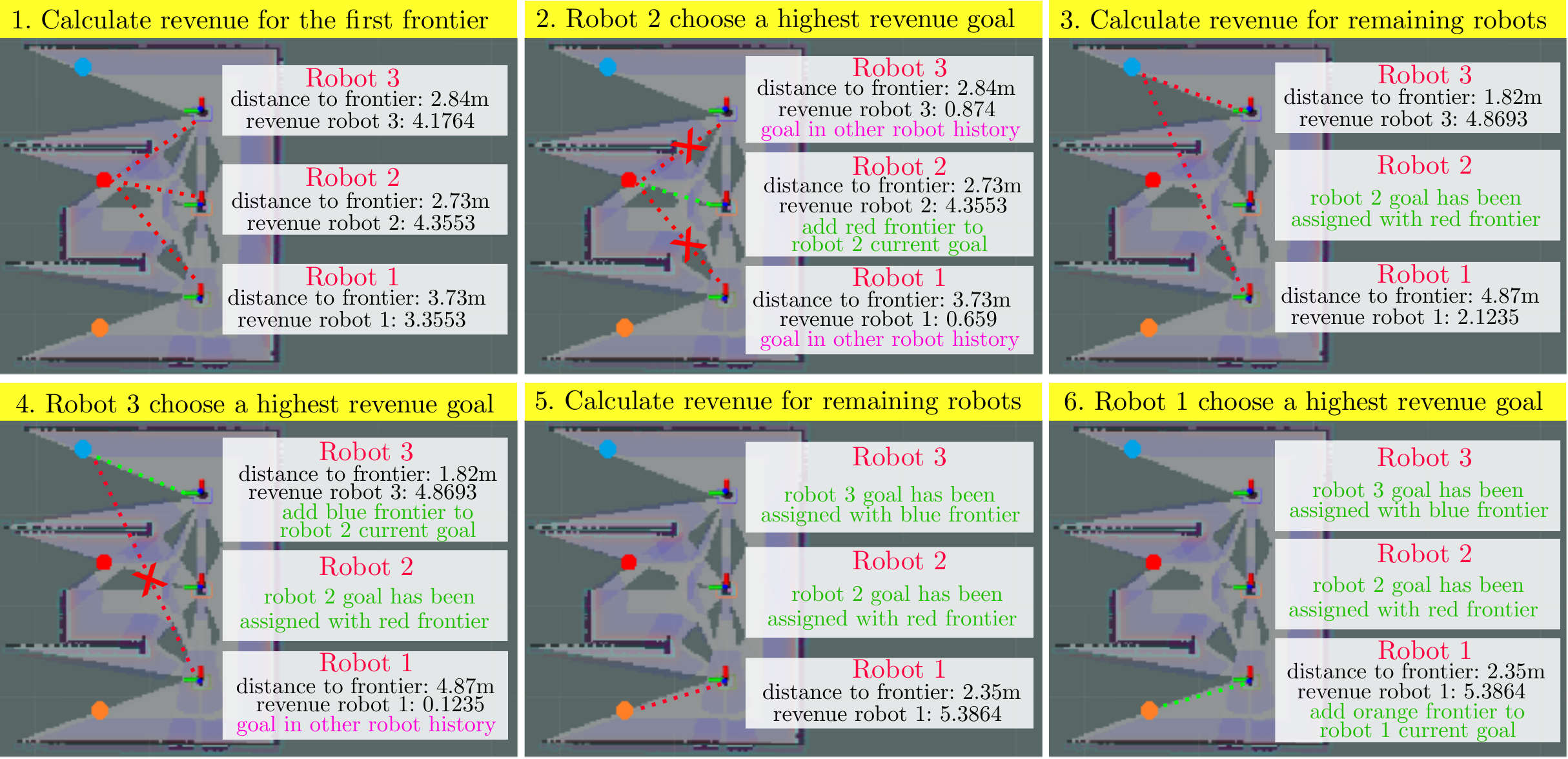} 
	\vspace{-0.6cm}
	\caption{Illustration of revenue calculation of TM-RRT exploration for three AGVs using Algorithm~\ref{alg:rrtAssign}, which RRT exploration would only consider the same frontier with highest revenue for all AGVs.}
	\vspace{-0.26cm}
	\label{fig:example revenue}
	\vspace{-0.16cm}
\end{figure}
If the frontier passes through all three conditions, the frontier will be sent to the AGV as goal and the AGV's goal history will be updated accordingly.
The revenue function of TM-RRT exploration can be found in the following Fig.~\ref{fig:example revenue}.
In the same computing cycle, the RRT exploration would assign the same goal to all the robots available given the frontier has the highest revenue. 
In contrast, TM-RRT exploration would calculate the revenue based on $RP_{dist}$ for each goal assigning cycle and prevents other AGVs from assigning the same goal.

\section{Simulation}
\label{sec:simulation}

\begin{figure}[h]
	\centering
	\includegraphics[width=0.431\textwidth]{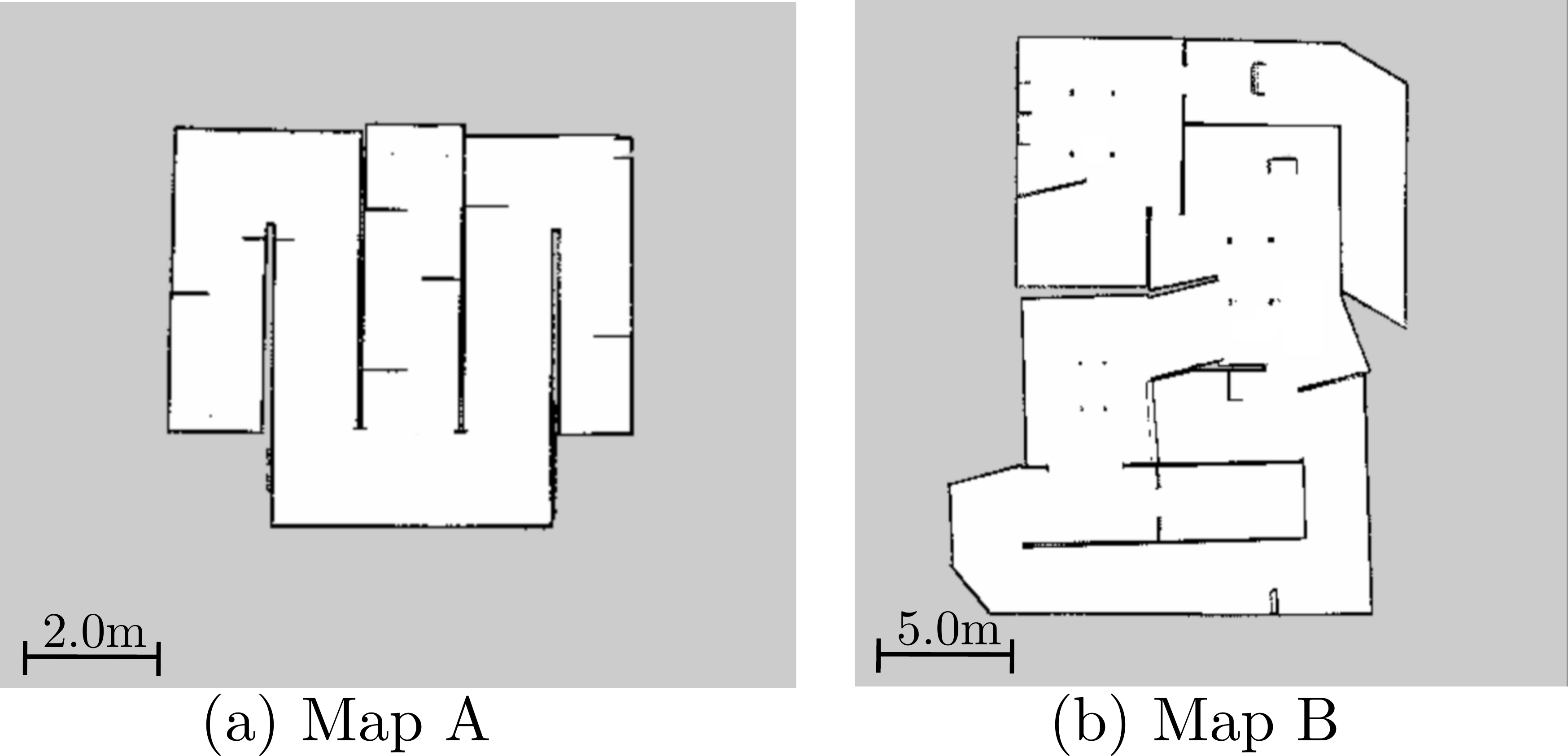} 
	\vspace{-0.2cm}
	\caption{Map A is a simple map with three path ways, while Map B is a complex map with dead-end and long stretch of corridor.}
	\label{fig:exp_map_merged}
	\vspace{-0.46cm}
\end{figure}

In this section, we conduct simulations on two aspects, which are studying TM-RRT exploration performance in different complexity of map and examining the ideal $rp_{dist}$ parameter.
Note that the simulation is conducted on Ubuntu 18.04 workstation and ROS Melodic with the specification as follows: Core i9 10900K, 64GB ram, and Nvidia RTX A6000.
The number of AGV used for the simulation of TM-Exploration RRT is three Turtlebot Waffle Pi and equipped with sensors such as 2D LiDAR, IMU, and RGB camera.
The localization of the AGV is done using Gmapping~\cite{Abdelrasoul2016quantitative}, while the navigation stack uses TEB planner~\cite{Roesmann2017Kinodynamic}.
The maps we used in the simulation are illustrated in Fig.~\ref{fig:exp_map_merged}, which map A is a three-way simple map and map B is a map with complex features such as a long corridor with a long dead-end.
Total surface area of map A is $120.0m^2$($10.0m\times12.0m$), while map B is $376.25m^2$($17.5m\times21.5m$).
Note that the simulation is visualized using RVIZ.

\vspace{-0.16cm}

\subsection{Benchmark and Comparison}
\label{subsec:benchmark}
In this subsection, we compare the performance of RRT exploration~\cite{Umari2017Autonomous} against TM-RRT exploration in two different environments, which one is simple and another is complex.
The first environment we used to compare both exploration methods is a simple map (Map A) with three different paths as shown in Fig.~\ref{fig:exp_map_merged}a. 
The map has a total area of $120.0m^2$, where three AGVs are spawned at the same initial location in the cross-section.
The purpose of performing exploration in this kind of environment is to study the efficiency of TM-RRT exploration handling a simple environment.

The parameters used in this simulation are listed as follows: [$rp_{dist}=18.0m$, $k=8.0$, $h=3.0$]. 
Note that parameters such as hysteresis gain and adaptive time are constant throughout the simulation.
For each exploration algorithm, the simulation is iterated $25$ times and we evaluate the performance using these two metrics, which are exploration duration and traveled distance. 
	The simulation of different metrics are presented in following Fig.~\ref{fig:simulation_result}.

\begin{figure}[h]
	\centering
	\includegraphics[width=0.49\textwidth]{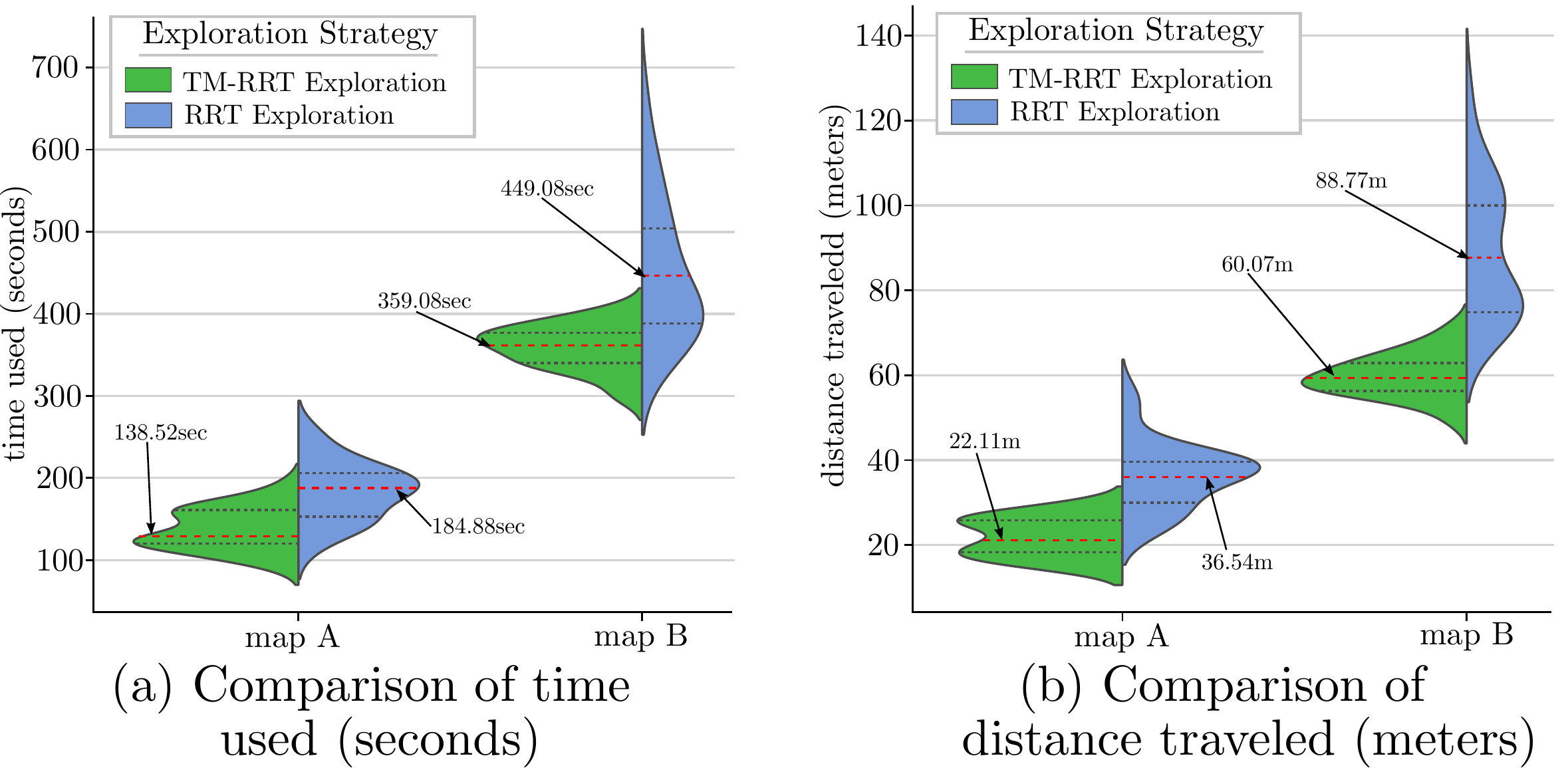} 
	\vspace{-0.6cm}
	\caption{Violin plot for TM-RRT and RRT exploration based on two metrics: time and distance traveled. Note that red color dotted line represents mean value of each metric.}
	\label{fig:simulation_result}
	\vspace{-0.45cm}
\end{figure}

Based on the Fig.~\ref{fig:simulation_result}'s result on map A, the TM-RRT exploration has a shorter exploration duration and traveled distance compared to RRT exploration. 
In terms of the average explored duration, TM-RRT~\cite{Umari2017Autonomous} explores the unknown region up to $33.46\%$($184.88$sec vs $138.52$sec) faster and traveled around $65.26\%$($36.54m$ vs $22.11m$) lesser.
The RRT exploration tends to have two or more AGVs exploring the same region due to the similar revenue calculated for the frontier detected.
The TM-RRT exploration accounts for the $rp_{dist}$ distance and therefore revenue for each AGV will not be solely based on the navigation cost and information gain. 
After investigating Map A, we move on to a more complex map, which is Map B. 

From Fig.~\ref{fig:simulation_result} in map B, the TM-RRT exploration on average outperforms the RRT exploration by $24.76\%$ ($449.08$sec vs $359.96$sec) in exploration duration and $47.78\%$ ($88.77m$ vs $60.07m$) in traveled distance.
In order to ensure each exploration strategy has 25 runs of exploration for calculating the average duration taken and distance traveled, we will rerun the exploration simulation if the current exploration process is stuck.
As observed, the RRT exploration~\cite{Umari2017Autonomous} has three attempts of exploration that are stuck, which it is unable to finish exploring the region.
Hence, we have performed an additional three more simulations for RRT exploration resulting in $28$ runs in total when compared to TM-RRT exploration.
In contrast, if an AGV stuck in a goal for too long, TM-RRT exploration can mark unreachable frontier as invalid and continue to explore.
The integration with the move base module for each AGV allows the TM-RRT exploration to monitor each robot state and reassign a new frontier if the current frontier is unreachable or the AGV is stuck.

\subsection{Exploring $rp_{dist}$ Parameter}
\label{subsec:relativeDistanceExp}
\begin{figure}[h]
	\vspace{-0.1cm}
	\centering
	\includegraphics[width=0.48\textwidth]{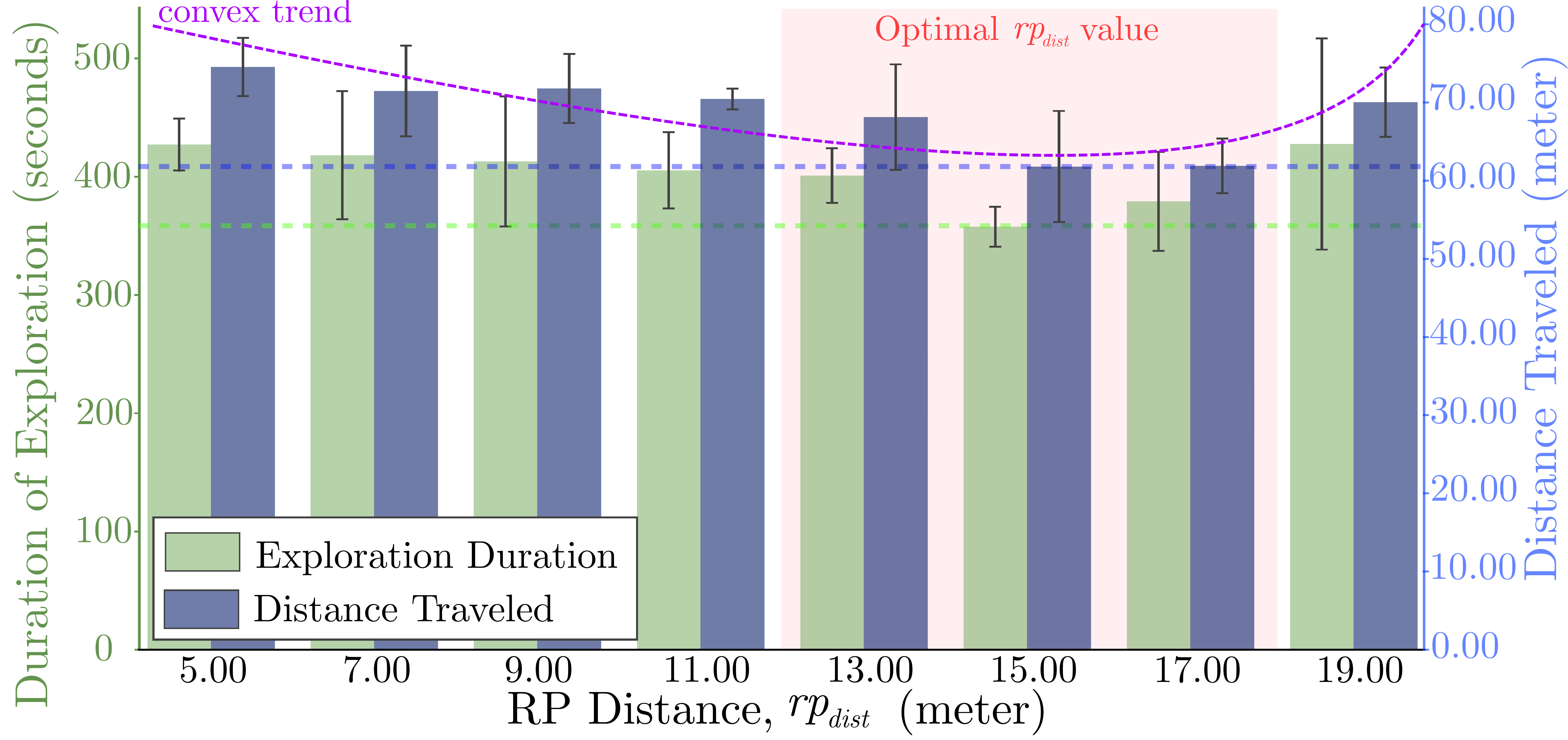} 
	\vspace{-0.2cm}
	\caption{Simulation of $rp_{dist}$ parameter against exploration duration and distance traveled.}
	\label{fig:exampleHybridExtraction}
	\vspace{-0.46cm}
\end{figure}

To study the $rp_{dist}$ parameter effect on the exploration efficiency, we choose map B as shown in Fig~\ref{fig:exp_map_merged}(b) to perform this study.
The $rp_{dist}$ distance parameter used in this simulation is set to a range of $5.0m$ to $9.0m$, where other parameters are fixed at [$k=8.0$, $h=3.0$].
Each parameter is studied with the simulation running for $5$ iterations. 
We use the same performance metric from the previous simulation to evaluate the exploration performance.
The result of the simulation is presented in Fig.~\ref{fig:exampleHybridExtraction}.

Based on the result observation, we notice the $rp_{dist}$ parameter performs better when $rp_{dist}$ distance is around $13.0m$, $15.0m$, and $17.0m$, as shown in the red highlighted box.
However, when the $rp_{dist}$ value is lower than $13.0m$, the distance traveled and exploration is slightly higher. 
This is because AGVs tend to move towards the same direction, resulting in more overlap in the traveled path, and thus resulting a longer exploration duration and higher traveled distance. 
On contrary, when the AGVs explore with a higher $rp_{dist}$ value, it tends to perform less ideal.
The main reason is the AGVs mostly treat each frontier with discounted revenue and less active exploring frontier with lower revenue. 
Hence, it reduces the overall exploration performance if the $rp_{dist}$ is not configured properly.
After studying the $rp_{dist}$ distance, the importance of selecting $rp_{dist}$ distance is crucial for optimal exploration performance. 
This could be further studied as an optimization problem as the result slightly resembles a convex curve, which is illustrated as purple line in Fig.~\ref{fig:exampleHybridExtraction}.

\vspace{-0.25cm}
\section{Real-world Deployment}
\label{sec:deploymnet}
\vspace{-0.07cm}
After testing TM-RRT exploration effectiveness in the simulation, we deploy it in multi-robot exploration using three four-wheeled AGV - Clearpath's Jackal. 
The Clearpath's Jackal is equipped with sensors such as Intel Realsense T265 and Velodyne VLP16.
The setup of the sensors and environment are shown in the following Fig.~\ref{fig:jackal_A_environment}(a).
\begin{figure}[h]
	\vspace{-0.1cm}
	\centering
	\includegraphics[width=0.48\textwidth]{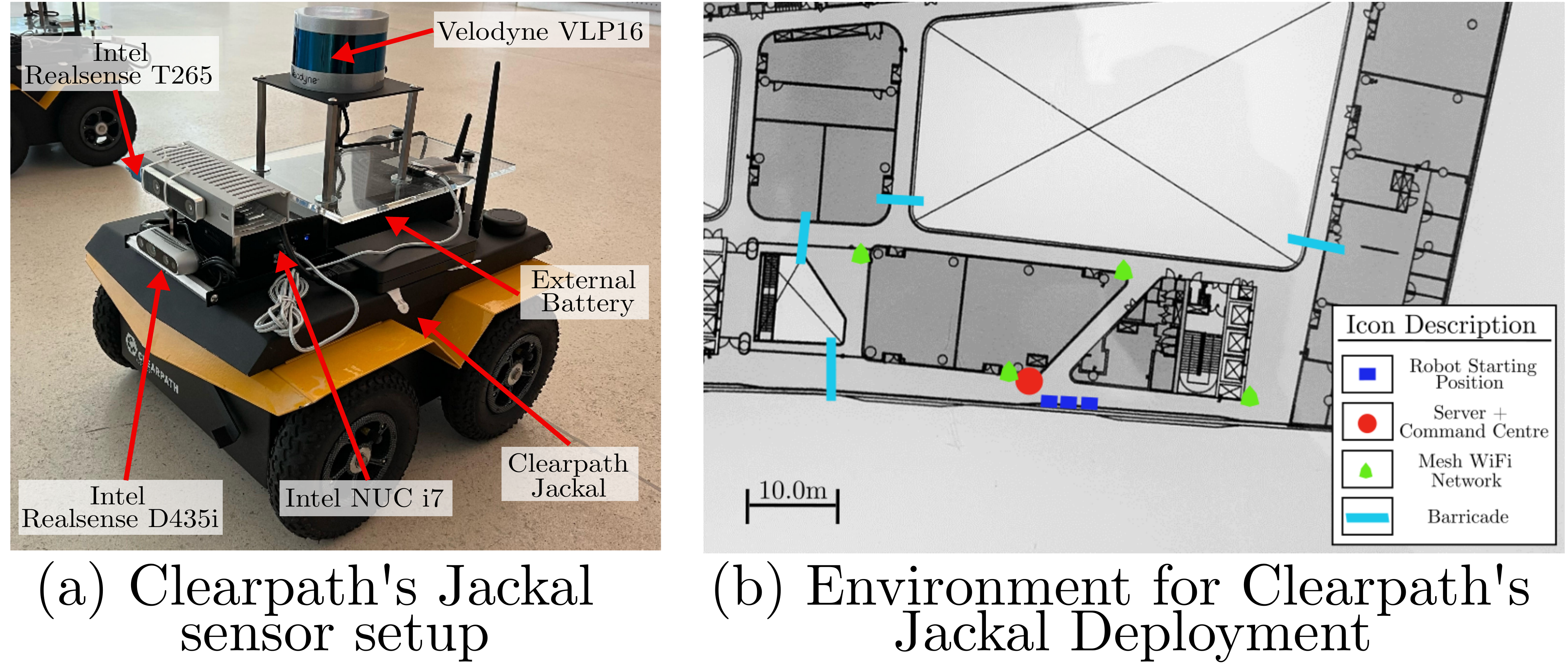} 
	\vspace{-0.2cm}
	\caption{Clearpath's Jackal sensor setup and deployment environment.}
	\label{fig:jackal_A_environment}
	\vspace{-0.34cm}
\end{figure}

The Jackal's computing unit for processing localization and navigation is done using Intel NUC i7 and it is connected to the jackal processing unit using an RJ45 cable.
The main connection between Jackal and server utilize WiFi 2.4GhZ mesh network to cover experiment's environment as presented in Fig.~\ref{fig:jackal_A_environment}(b).
Four mesh network routers(Asus RT-AX58U) are deployed to provide enough coverage for the network, and each Jackal has a custom script running to switch the network if needed.
The total surface area of experiment's environment is roughly $1350.0m^2$ area, and barricades are used to block off-pathway to prevent exploring region that is not intended for the experiment.

\begin{figure}[h]
	\centering
	\includegraphics[width=0.44\textwidth]{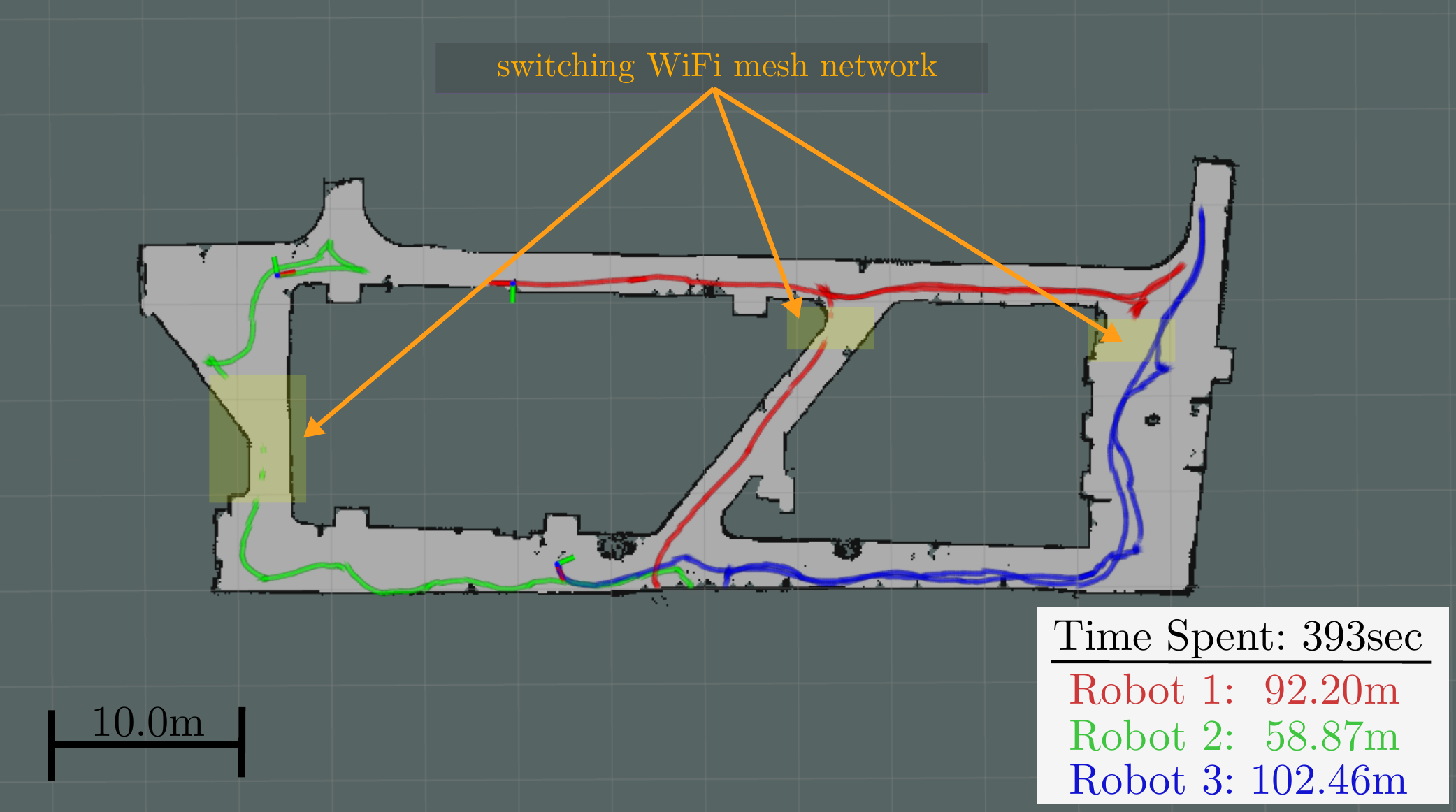} 
	\vspace{-0.2cm}
	\caption{Exploration with three Clearpath's Jackals. Red, Green, and Blue represents each Jackal's traveled path. Note that there is WiFi switching causes some missing odometry data due to data loss as shown in highlighted yellow color.}
	\label{fig:final_map_jackal}
	\vspace{-0.34cm}
\end{figure}

We use Gmapping~\cite{Abdelrasoul2016quantitative} for each Jackal's localization and mapping.
To adapt bigger environment, the map resolution used for merging has been adjusted to $0.15$ and the map merging package's frequency has been adjusted to $0.1$hz for optimal communication efficiency.
We use the following parameters for the TM-RRT exploration: [$rp_{dist}=25.0m$, $k=6.0$, $h=5.0$]
The outcome of the exploration is shown in Fig.~\ref{fig:final_map_jackal}.

The visualization of each jackal's traveled path is distinguished using three different colors, which are red, green, and blue representing Jackal $1$, Jackal $2$, and Jackal $3$ respectively.
The whole exploration process took around $6$mins $33$sec and each AGV traveled $92.20m$, $58.87m$, and $102.46m$, resulting in an average traveled distance of $84.51m$.
The implementation of the TM-RRT exploration has shown successful attempts in exploring the large unknown region.
Also, it did help the Jackals to spread out themselves while exploring the unknown region.
We try to implement RRT exploration as well but it does not perform well due to AGV frequently being stuck at unreachable goals.

\section{Discussion and Conclusion}
\label{sec:conclusion}
In this paper, we present TM-RRT exploration for multiple AGVs in an unknown environment. 
In addition, we also propose a map post-processing method that helps with reducing localization noises from larger LiDAR scanning range.
The TM-RRT exploration leverages a temporal-based mission assignment to designate a time limit for each frontier assigned to AGV.
This prevents AGV from spending too much time exploring the same frontier or being stuck in an unreachable frontier. 
Another important element of the TM-RRT exploration is keeping track of the AGV's goal history, which avoid AGV having the same frontier goal assigned to different AGV.
To further improve the goal assignment, revenue functions also use AGV's relative position when calculating the revenue for each frontier detected. 
The proposed exploration has been tested in simulation and real-world deployment. 
Results have shown it performs better than conventional RRT exploration and it is faster in terms of exploration duration ($33.46\%, 24.76\%$) and traveled distance ($65.26\%,47.78\%$).

In future work, the TM-RRT exploration strategy can be further improved in a few directions. 
Goal assignment can be further optimized with deep neural network methods. 
For instance, the input of frontiers and map can be used as input, which later output goal assignment for the number AGVs present in the exploration.
The optimization of the $rp_{dist}$ to adapt to a different environment can be also an interesting topic to explore in future work.
Additionally, additional experiments can be carried out to ensure the robustness of the TM-RRT exploration.
The next research focus would be upgrading the centralized paradigm to a distributed architecture.
A major limitation of the centralized approach is that AGV would be required to be connected to the central for goal assignment, which requires high reliability in the communication. 
On the other hand, a distributed approach offloads the goal assigned to the individual AGV.
This would be more robust against connection loss and operate independently of other AGVs. 
Additionally, the study of implementing TM-RRT exploration in a heterogeneous robot system or infrastructure-free environment is another potential directions to enhance exploration strategy.

\bibliographystyle{IEEEtran}
\bibliography{bibSpace}

%

%
%
%




\end{document}